\documentclass[]{llncs}
\usepackage{graphicx}
\usepackage{subcaption}
\usepackage{xcolor}
\usepackage{adjustbox}
\usepackage{array}

\usepackage{epsfig}
\usepackage{graphicx}

\usepackage{hyperref}
\hypersetup{colorlinks=true}

\graphicspath{{./figures/}}
\usepackage{amsmath}
\usepackage{lineno}
\usepackage{bm}
\usepackage[draft]{fixme}
\usepackage{xcolor}
\usepackage{booktabs, wrapfig}
\usepackage{multirow}

\newcommand{\ie}{\textit{i}.\textit{e}.}
\newcommand{\eg}{\textit{e}.\textit{g}.}

\newcommand{\Eg}{\textit{E}.\textit{g}.}

\begin{document}
\title{DeepAtlas: Joint Semi-Supervised Learning of Image Registration and Segmentation}

%
\author{Zhenlin Xu \and Marc Niethammer}

%
%
\institute{University of North Carolina, Chapel Hill, NC, USA}
%
\maketitle              
\begin{abstract}
Deep convolutional neural networks (CNNs) are state-of-the-art for semantic image segmentation, but typically require many labeled training samples. Obtaining 3D segmentations of medical images for supervised training is difficult and labor intensive. Motivated by classical approaches for joint segmentation and registration we therefore propose a deep learning framework that jointly learns networks for image registration and image segmentation. In contrast to previous work on deep unsupervised image registration, which showed the benefit of weak supervision via image segmentations, our approach can use existing segmentations when available and computes them via the segmentation network otherwise, thereby providing the same registration benefit. Conversely, segmentation network training benefits from the registration, which essentially provides a realistic form of data augmentation. Experiments on knee and brain 3D magnetic resonance (MR) images show that our approach achieves large simultaneous improvements of segmentation and registration accuracy (over independently trained networks) and allows training high-quality models with very limited training data. Specifically, in a one-shot-scenario (with only one manually labeled image) our approach increases Dice scores (\%) over an unsupervised registration network by 2.7 and 1.8 on the knee and brain images respectively. 
\end{abstract}

\section{Introduction}
\label{section:introduction}
Image segmentation and registration are two crucial tasks in medical image analysis. They are also highly related and can help each other. \Eg, labeled atlas images are used via image registration for segmentation. Segmentations can also provide extra supervision (in addition to image intensities) for image registration and are used to evaluate registration results.
Consequentially, joint image registration and segmentation approaches have been proposed. \Eg, approaches based on active-contours~\cite{jointContour2001} and Bayesian~\cite{pohl2006bayesian} or Markov random field formulations~\cite{mahapatra2010joint}. While these methods jointly estimate registration and segmentation results, they operate on \emph{individual} image pairs (instead of a population of images) and require the computationally costly minimization of an energy function. 

Deep learning (DL) has been widely and successfully applied to medical image analysis. For supervised image segmentation, CNN-based approaches are faster and better than classical methods when many labeled training samples are available~\cite{litjens2017survey}. DL-based registration achieves similar performance to optimization-based approaches but is much faster. As true transformations are not available, training either uses estimates from optimization-based methods~\cite{yang2017quicksilver} or is unsupervised~\cite{balakrishnan2018unsupervised}. Recent work~\cite{balakrishnan2019voxelmorph} shows that weak supervision via an additional image segmentation loss between registered images can improve  results over unsupervised training, which relies on the images alone. In practice, obtaining segmentations for 3D medical images is difficult and labor intensive. Hence, manual segmentations will often not be available for a large fraction of image data.
 
We propose \textbf{DeepAtlas}, to jointly learn deep networks for weakly supervised registration and semi-supervised segmentation.
Our contributions are:
\begin{itemize}
	\item[$\bullet$] \emph{We propose the first approach to jointly learn two deep neural networks for image registration and segmentation}. Previous joint approaches require joint optimizations for each image pair. Instead, we jointly learn from a population of images during training, but can independently use the resulting segmentation and registration networks at test time. 
	\item[$\bullet$] \emph{Our joint approach only requires few manual segmentations}. Our two networks mutually guide each other's training on unlabeled images via an anatomy similarity loss. This loss penalizes the dissimilarity of the warped segmentation of the moving image and the segmentation of the target image. When registering image pairs consisting of a manually labeled image and the estimate of a labeled image (via its network-predicted segmentation), this loss provides anatomy consistency supervision for registration and forces the predicted segmentation to match the manual segmentation after registration.  
    \item[$\bullet$] \emph{We evaluate our approach on large 3D brain and knee MRI datasets}. Using few manual segmentations, our method outperforms separately learned registration and segmentation networks. In the extreme case, where only one manually segmented image is available, our approach facilitates one-shot segmentation and boosts registration performance at the same time. 
\end{itemize}

\section{Method}
\label{section:method}
Our goal is to improve registration and segmentation accuracy when few manual segmentations are available for a large set of images by jointly learning a segmentation and a registration network. Fig.~\ref{fig:diagram} illustrates our approach consisting of two parts: weakly-supervised registration learning (solid blue lines) and semi-supervised segmentation learning (dashed yellow lines). Our loss is the weighted sum of the registration regularization loss ($\mathcal{L}_{r}$), the image similarity loss ($\mathcal{L}_{i}$), the anatomy loss ($\mathcal{L}_{a}$) penalizing segmentation dissimilarity, and the supervised segmentation loss ($\mathcal{L}_{sp}$).
The losses \{$\mathcal{L}_{r}$, $\mathcal{L}_{i}$, $\mathcal{L}_{a}$\} drive the weakly supervised learning of registration (Sec.~\ref{subsec:registration}) and the losses \{$\mathcal{L}_{a}$, $\mathcal{L}_{sp}$\} drive the semi-supervised learning of segmentation (Sec.~\ref{subsec:segmentation}). Sec.~\ref{subsec:implementation} details the implementation.

\subsection{Weakly-supervised Registration Learning}
\label{subsec:registration}
Given a pair of moving and target images $I_m$ and $I_t$, a registration network $\mathcal{F}_R$ with parameters $\theta_r$ predicts a displacement field $\mathbf{u} = \mathcal{F}_R(I_m, I_t; \theta_r)$. This then allows warping the moving image to the target image space, $I_m^w = I_m \circ \Phi^{-1}$, where $\Phi^{-1} = \mathbf{u} + {\rm id}$ is the deformation map and ${\rm id}$ is the identity transform. A good map, $\Phi$, maps related anatomical positions to each other.
\begin{figure}[tb]
	\begin{center}
		\includegraphics[width=\textwidth]{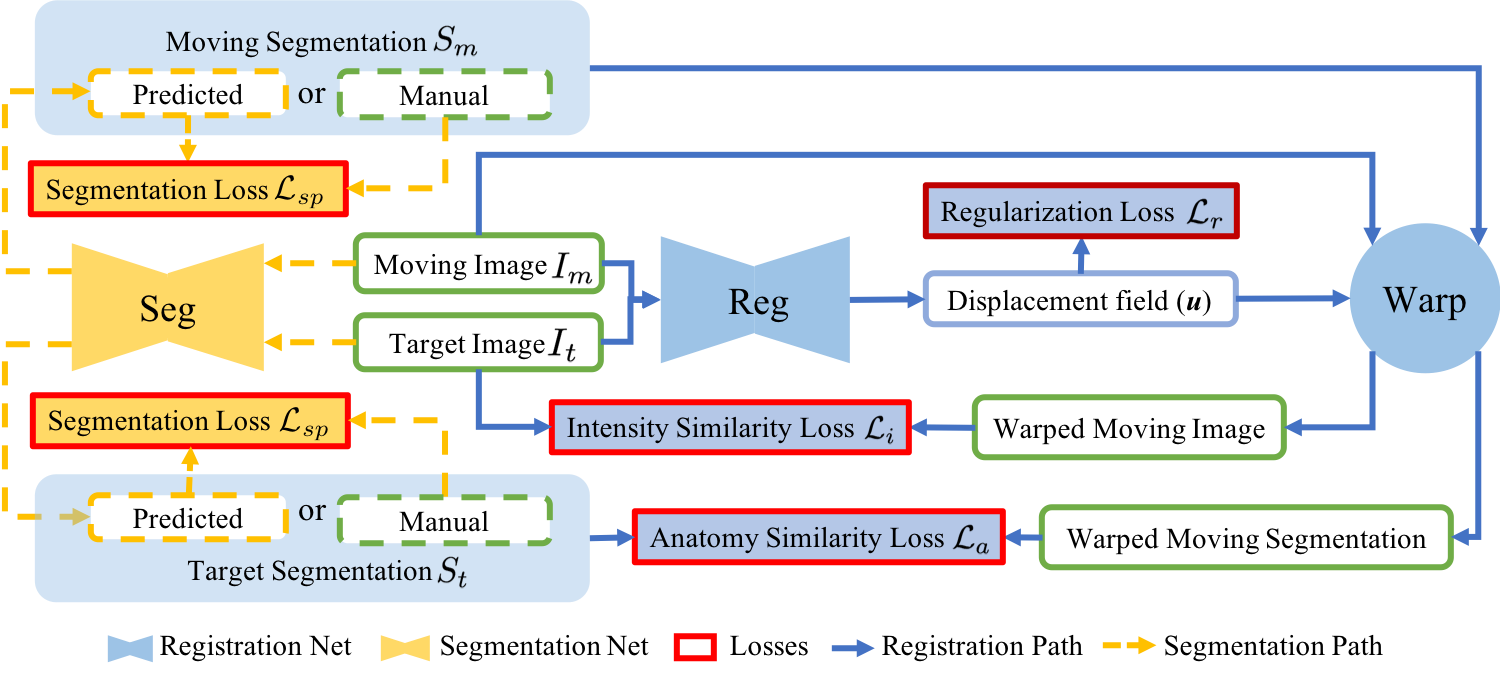}	
	\end{center}
	\caption{\emph{DeepAtlas} for joint learning of weakly supervised registration and semi-supervised segmentation. Unlabeled moving/target images are segmented by the segmentation network so that every training registration pair has weak supervision via the anatomy similarity loss which also guides segmentation learning on unlabeled images.}
	\label{fig:diagram}
\end{figure}
Unsupervised registration learning optimizes $\theta_r$ over an intensity similarity loss $\mathcal{L}_{i}$ (penalizing appearance differences between $I_t$ and $I_m^w$) and a regularization loss $\mathcal{L}_{r}$ on $\mathbf{u}$ to encourage smooth transformations. Adding weak supervision by also matching segmentations between the target image ($S_t$) and the warped moving image ($S_m^w=S_m\circ\Phi^{-1}$) via an anatomy similarity loss $\mathcal{L}_{a}$ can improve registrations~\cite{balakrishnan2019voxelmorph}. Weakly-supervised registration learning is then formulated as:
\begin{align}
\label{eq:reg_learning}
\theta_r^\star = \underset{\theta_r}{\rm argmin}\{
  \mathcal{L}_{i}(I_m \circ \Phi^{-1}, I_t) + \lambda_{r} \mathcal{L}_{r}(\Phi^{-1}) + \lambda_{a}\mathcal{L}_{a}(S_m \circ\Phi^{-1}, S_t)\},
\end{align}
with weights $\lambda_r, \lambda_a\geq 0$. 
In practice, while a large set of images are often available, few of them have manual segmentations. In contrast to existing work, we estimate missing moving or target segmentations via our segmentation network (see Fig.~\ref{fig:diagram}). Hence, we provide weak supervision for \emph{every} training image pair.

\subsection{Semi-supervised Segmentation Learning}
\label{subsec:segmentation}
The segmentation network $\mathcal{F}_{S}$ with parameters $\theta_s$ takes an image $I$ as input and generates probabilistic segmentation maps for all semantic classes: $\hat{S} = \mathcal{F}_{S}(I;\theta_s)$. 
In addition to the typical supervised segmentation loss $\mathcal{L}_{sp}(\hat{S},S)$ where $S$ is a given manual segmentation, the anatomy similarity loss for registration $\mathcal{L}_{a}(S_m \circ\Phi^{-1}, S_t)$ also drives segmentation learning when $S_m$ or $S_t$ are predicted via $\mathcal{F}_{S}$ for unlabeled images.
Specifically, we define these losses as:
\begin{equation*}
\label{eq:seg_losses}
\mathcal{L}_{seg} = 
\begin{cases}
 \lambda_{a}\mathcal{L}_{a}(S_m \circ\Phi^{-1}, \mathcal{F_S}(I_t)) 
+ \lambda_{sp}\mathcal{L}_{sp}(\mathcal{F_S}(I_m), S_m),  \text{ if $I_t$ is unlabeled;}\\   
\lambda_{a}\mathcal{L}_{a}(\mathcal{F_S}(I_m) \circ\Phi^{-1}, S_t) 
+ \lambda_{sp}\mathcal{L}_{sp}( \mathcal{F_S}(I_t),S_t), \text{ if $I_m$ is unlabeled;}\\
\lambda_{a}\mathcal{L}_{a}(S_m \circ\Phi^{-1}, S_t) 
+ \lambda_{sp}\mathcal{L}_{sp}( \mathcal{F_S}(I_m),S_m), \text{if $I_m$ and $I_t$ are labeled;}\\
0,  \text{ if both $I_t$ and $I_m$ are unlabeled.}
\end{cases}
\end{equation*}
with weights $\lambda_a,\lambda_{sp}\geq 0$.
$\mathcal{L}_{a}$ teaches $\mathcal{F}_{S}$ to segment an unlabeled image such that the predicted segmentation matches the manual segmentation of a labeled image via $\mathcal{F}_{R}$. In the case where the target image $I_t$ is unlabeled, $\mathcal{L}_{a}$ is equivalent to a supervised segmentation loss on $I_t$, in which the single-atlas segmentation $S_m \circ\Phi^{-1}$ is the noisy true label. Note that we do not use two unlabeled images for training and $\mathcal{L}_a$ does not train the segmentation network when both images are labeled. We then train our segmentation network in a semi-supervised manner as follows:
\begin{equation}
\label{eq:seg_learning}
	\theta_s^\star = \underset{\theta_s}{\rm argmin}~ \mathcal{L}_{seg}.
\end{equation}

\subsection{Implementation Details}
\label{subsec:implementation}
\textbf{Losses:} Various choices are possible for the intensity/anatomy similarity, the segmentation, and the regularization losses. Our choices are as follows. 

\noindent \textit{Anatomy similarity and supervised segmentation loss:} A cross-entropy loss requires manually tuned class weights for imbalanced multi-class segmentations~\cite{ronneberger2015u}. We use a soft multi-class Dice loss which addresses imbalances inherently: 
\begin{equation}
\label{eq:dice_loss}
\mathcal{L}_{dice}(S,S^\star) = 1 - \frac{1}{K}\sum_{k=1}^{K} \frac{\sum_{x} S_k(x) S^\star_k(x)}{\sum_{x}S_k(x) + \sum_{x} S^\star_k(x)},
\end{equation}
where $k$ indicates a segmentation label (out of $K$) and $x$ is voxel location. $S$ and $S^\star$ are two segmentations to be compared.

\noindent \textit{Intensity similarity loss:} We use normalized cross correlation (NCC) as:
\begin{equation}
\mathcal{L}_{i}(I_m^w, I_t) = 1 - NCC(I_m^w, I_t),
\end{equation}

which will be in $[0,2]$ and hence will encourage maximal correlation.

\noindent \textit{Regularization loss:} We use the bending energy~\cite{Rueckert1999}:
 
\begin{equation}
	\mathcal{L}_{r}(\mathbf{u}) = \frac{1}{N}\sum_{\mathbf{x}}\sum_{i=1}^{d} \|H(u_i(\mathbf{x}))\|_F^2
\end{equation}
where $\|\cdot\|_F$ denotes the Frobenius norm, $H(u_i(\mathbf{x}))$ is the Hessian of the i-th component of $\mathbf{u(\mathbf{x})}$, and $d$ denotes the spatial dimension ($d=3$ in our case). $N$ denotes the number of voxels. Note that this is a second-order generalization of diffusion regularization, where one penalizes $\|\nabla u_i(\mathbf{x})\|_2^2$ instead of $\|H(u_i(\mathbf{x}))\|_F^2$.

\textbf{Alternating training:} It is in principle straightforward to optimize two networks according to Eqs.~\ref{eq:reg_learning} and~\ref{eq:seg_learning}. However, as we work with the whole 3D images, not cropped patches, GPU memory is insufficient to simultaneously optimize the two networks in one forward pass. Hence, we alternately train one of the two networks while keeping the other fixed. 
We use a 1:20 ratio between training steps for the segmentation and registration networks, as the segmentation network converges faster. Since it is difficult to jointly train from scratch with unlabeled images, we independently pretrain both networks. When only few manual segmentations are available, \eg, only one, separately training the segmentation network is challenging. In this case, we train the segmentation network from scratch using a fixed registration network trained unsupervisedly. We start alternating training when the segmentation network achieves reasonable performance. 

\textbf{Networks:} DeepAtlas can use any CNN architecture for registration and segmentation. We use the network design of~\cite{balakrishnan2018unsupervised} for registration; and a customized light 3D U-Net design for segmentation with LeakyReLU instead of ReLU, and smaller feature size due to GPU memory limitations. 

\section{Experiments and Results}
\label{section:experiments}
We show on a 3D knee and a 3D brain MRI dataset that our framework improves both registration and segmentation when many images with few manual segmentations are available: i.e. $N$ of $M$ images are labeled ($N<<M$).

\textbf{Mono-networks}: 
We train single segmentation/registration models as baselines. For segmentation, fully supervised networks are trained with $N$ labeled images; the registration networks are trained via Eq.~\ref{eq:reg_learning} using all $M$ training images with $N$ images labeled; the anatomy similarity loss, $\mathcal{L}_a$, is only used for training pairs where both images have manual segmentations. Models trained with $N=M$ manual segmentations (\ie, with manual segmentations for all images) provide our upper performance bound. All mono-networks are trained for a sufficient number of epochs until they over-fit. The best models based on validation performance are evaluated.

\textbf{DeepAtlas (DA)}: We initialize the joint model with the trained mono-networks. In addition to the alternately trained DA models, we hold one network fixed all through training, termed \textbf{Semi-DeepAtlas (Semi-DA)}. 

In one-shot learning (N=1) experiments, training a supervised segmentation network based on a single labeled image is difficult; hence, we do not compute a segmentation mono-network in this case. For Semi-DA, we train a segmentation network from scratch with a fixed registration network that is trained unsupervised (N=0). The DA model is initialized using the Semi-DA segmentation network and the unsupervised registration network.
\begin{figure} [tp]
\begin{center}
\renewcommand{\arraystretch}{0}
\newcommand\cwidth{0.14\textwidth}
\begin{adjustbox}{max width=\textwidth}
\begin{tabular}{ccccccc}
	Moving & Target &  Mono-0 & Mono-5 & DA-1 & DA-5 & Mono-200 {\smallskip}\\
\includegraphics[width=\cwidth]{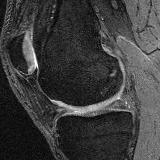} &
\includegraphics[width=\cwidth]{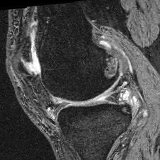} &
\includegraphics[width=\cwidth]{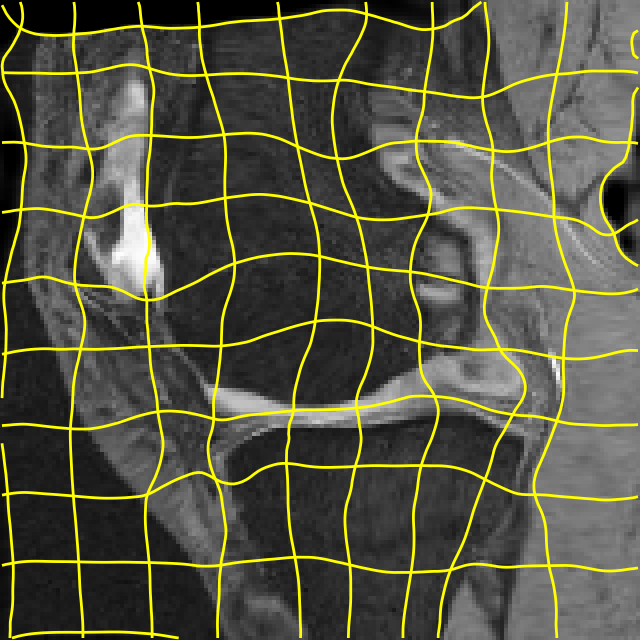} &
\includegraphics[width=\cwidth]{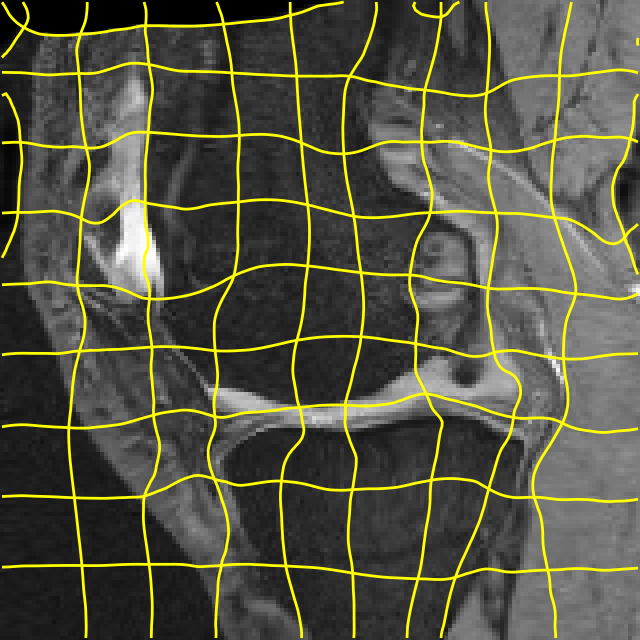} &
\includegraphics[width=\cwidth]{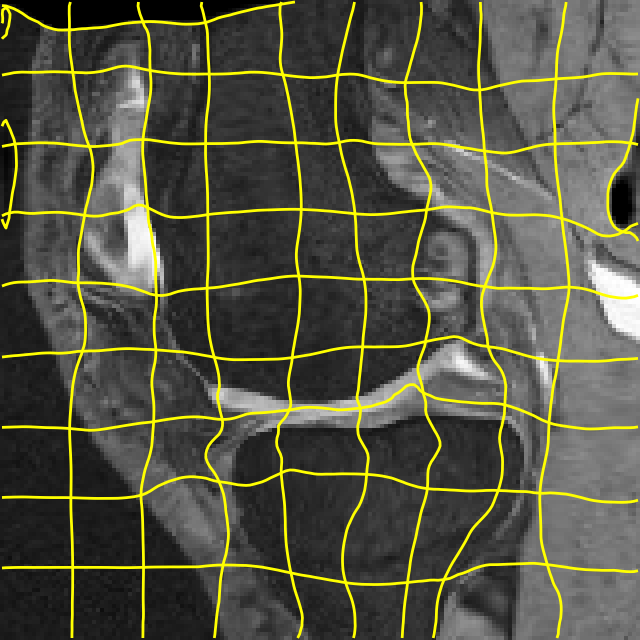} &
\includegraphics[width=\cwidth]{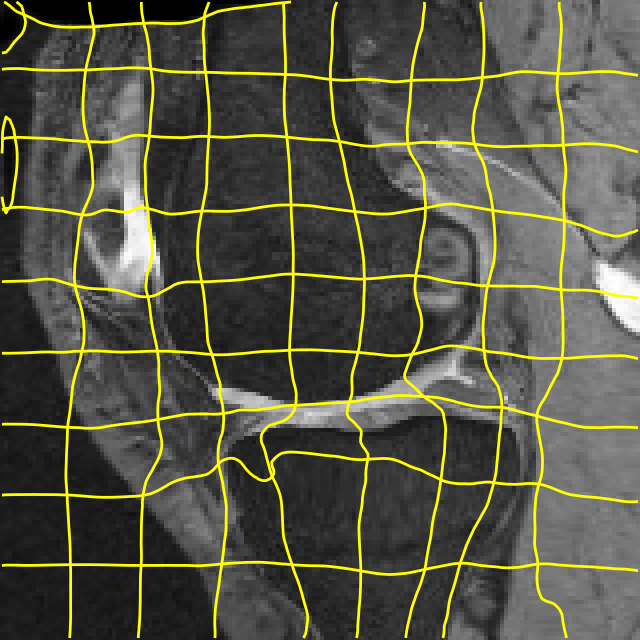} &
\includegraphics[width=\cwidth]{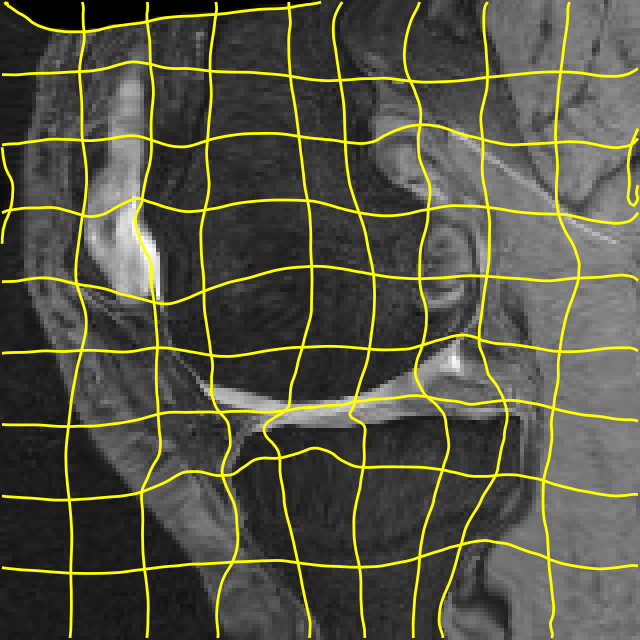} 
\\
\includegraphics[width=\cwidth]{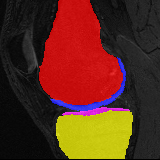} &
\includegraphics[width=\cwidth]{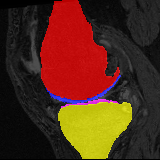} &
\includegraphics[width=\cwidth]{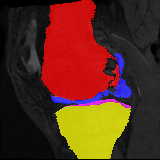} &
\includegraphics[width=\cwidth]{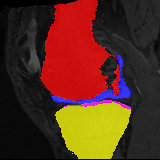} &
\includegraphics[width=\cwidth]{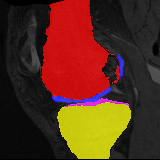} &
\includegraphics[width=\cwidth]{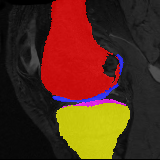} &
\includegraphics[width=\cwidth]{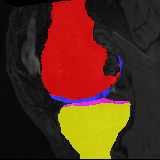} 
\end{tabular}
\end{adjustbox}
\end{center}

\begin{center}
\renewcommand{\arraystretch}{0}
\newcommand\cwidth{0.165\textwidth}
\begin{adjustbox}{max width=\textwidth}
\begin{tabular}{cccccc}
	Image & Manual Seg &  DA-1 & Mono-21 & DA-21 & Mono-65 {\smallskip}\\
\includegraphics[width=\cwidth]{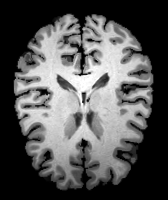} &
\includegraphics[width=\cwidth]{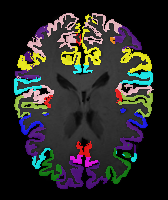} &
\includegraphics[width=\cwidth]{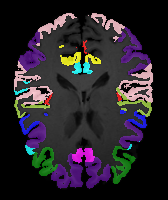} &
\includegraphics[width=\cwidth]{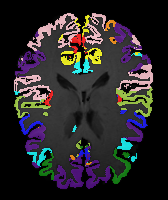} &
\includegraphics[width=\cwidth]{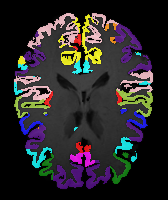} &
\includegraphics[width=\cwidth]{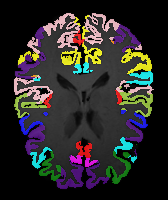} 
\\
\end{tabular}
\end{adjustbox}
\end{center}
\caption{\small Examples of knee MRI registration (top) and brain MRI segmentation (bottom) results. \textbf{Top}: The first two columns are the moving image/segmentation and the target image/segmentation followed by the warped moving images (with deformation grids)/segmentations by different models. \textbf{Bottom left to right}: original image, manual segmentation, and predictions of various models. Mono-$i$ and DA-$i$ represent the mono- and DA models with $i$ manual segmentations respectively.}
\label{fig:qualitative_results}
\end{figure}

\textbf{Knee MRI experiment:} We test our method on 3D knee MRIs from the Osteoarthritis Initiative (OAI)  \footnote{https://nda.nih.gov/oai/} 
and corresponding segmentations of femur and tibia as well as femoral and tibial cartilage~\cite{AmbellanTackEhlkeetal2019OAIZIB}. From a total of 507 labeled images, we use 200 for training, 53 for validation, and 254 for testing. To test registration performance we use 10,000 random image pairs from the test set. All images are affinely registered to an atlas built from the training images, resampled to isotropic spacing of 1mm, cropped to $160\times160\times160$ and intensity normalized to [0,1]. In addition, right knee images are flipped to be consistent with left knees. For training, the loss weights are $\lambda_{r}=20,000$, $\lambda_{a}=3$, and $\lambda_{sp}=3$ based on approximate hyper-parameter tuning. Note that when computing $\mathcal{L}_r$ from the displacements, the image coordinates are scaled to [-1, 1] for each dimension following the convention in the interpolation function of PyTorch.

\textbf{Brain MRI experiment:} We also evaluate our method on the MindBooggle101~\cite{MindBoggle101} brain MRIs with 32 cortical regions. We fuse corresponding segmentation labels of the left and right brain hemispheres. MindBoogle101 consists of images from multiple datasets, \eg, OASIS-TRT-20, MMRR-21 and HLN-12. After removing images with incorrect labels, we obtain a total of 85 images. We use 5 images from OASIS-TRT-20 as validation set and 15 as test set. We use the remaining 65 images for training. Manual segmentations in the N=1 and N=21 experiments are only from the MMRR-21 subset; this simulates a common practical use case, where we only have few manual segmentations for one dataset and additional unlabeled images from other datasets, but desire to process a different, new dataset. All images are 1mm isotropic, affinely-aligned, histogram-matched, and cropped to size $168\times200\times169$. We apply sagittal flipping for training data augmentation. We use the same loss weights as for the knee MRI experiment except for $\lambda_{r}=5,000$, since cross-subject brain registrations require large deformations and hence less regularization. 

\setlength{\tabcolsep}{1pt}
\begin{table*}[t]
\centering
\normalsize
\begin{adjustbox}{max width=\textwidth}
	\begin{tabular}{>{\centering\arraybackslash}m{2em}|c|ccc|ccc}
	\hline
	\multirow{2}{*}{$N$}  & \multirow{2}{*}{Models} & \multicolumn{3}{c|}{Segmentation Dice ($\%$)} & \multicolumn{3}{c}{Registration Dice ($\%$)} \\ 
	\cline{3-8}
	& & Bones & Cartilages & All & Bones & Cartilages & All \\
	\hline
	0 & Mono & - & - & - & 95.32(1.13) & 65.71(5.86) & 80.52(3.24)\\
	\hline
	
	\multirow{2}{*}{1} 	& {Semi-DA} &  96.43(0.85) & 76.67(3.24)& 86.55(1.86)  & - & - & -\\
						& {DA} & \textbf{96.80(0.81)} & \textbf{77.63(3.22)} & \textbf{87.21(1.84)} 
						& 95.76(1.01) & 70.77(5.68) & 83.27(3.14)\\
	\hline
	
	\multirow{3}{*}{5} 	
	& Mono &96.51(1.69) & 78.95(3.91) & 87.73(2.37)
	& 95.60(1.08) & 68.13(5.98) &  81.87(3.31)\\
	& {Semi-DA} & 96.97(1.26) & 79.73(3.84)  & 88.35(2.22)
	& \textbf{96.38(0.81)} & 73.48(5.26) & 84.93(2.89) \\
	& {DA} & \textbf{97.49(0.67)} & \textbf{80.35(3.64)} & \textbf{88.92(2.01)} 
	& 96.35(0.82) & \textbf{73.67(5.22)} & \textbf{85.01(2.86)}\\
	\hline
	\multirow{3}{*}{10} 	
	& Mono & 97.29(1.03) & 80.59(3.67) & 88.94(2.07)
	& 95.77(1.02) & 69.45(5.93)& 82.61(3.27)  \\
	& {Semi-DA} & 97.60(0.76) & \textbf{81.21(3.58)} & 89.40(1.99) 
	& \textbf{96.66(0.72)} & 74.67(5.01)  & \textbf{85.66(2.73)}\\
	& {DA} & \textbf{97.70(0.65)} & 81.19(3.47)  & \textbf{89.45(1.91)}
	&  96.62(0.75) & \textbf{74.69(5.03)} &  85.66(2.75)\\
	\hline
	200 & Mono & 98.24(0.34) & 83.54(2.93) & 90.89(1.56) & 96.98(0.56) & 77.33(4.34) & 87.16(2.35)\\ 
	\hline
	\end{tabular}
\end{adjustbox}
\caption{Segmentation and registration performance on 3D knee MRIs. Average (standard deviation) of Dice scores (\%) for bones (femur and tibia) and cartilages (femoral and tibial). N of 200 training images are manually labeled.}
\label{table:oai}
\end{table*}

\textbf{Optimizer:} We use \texttt{Adam}. The initial learning rates are 1e-3 for the mono-networks. Initial learning rates are 5e-4 for the registration network and 1e-4 for the segmentation network for Semi-DA and DA. Learning rates decay by 0.2 at various epochs across experiments. We use PyTorch and run on Nvidia V100 GPUs with 16GB memory.

\textbf{Results:} All trained networks are evaluated using Dice overlap scores between predictions and the manual segmentations for the segmentation network, or between the warped moving segmentations and the target segmentations for the registration network. Tabs.~\ref{table:oai} and \ref{table:mindboogle} show results for the knee and brain MRI experiments respectively in Dice scores (\%). Fig.~\ref{fig:qualitative_results} shows examples of knee MRI registrations and brain MRI segmentations. 

\emph{General results:} For both datasets across different numbers of manual segmentations, Semi-DA, which uses a fixed pre-trained network to help the training of the other network, boosts performance compared to separately trained mono-networks. DA, where both networks are alternately trained, achieves even better Dice scores in most cases. Based on a Mann-Whitney U-test with a significance level of 0.05 and a correction for multiple comparisons with a false discovery rate of 0.05, our models (DA/Semi-DA) result in significantly larger Dice scores than the mono-networks for all experiments. This demonstrates that segmentation and registration networks can indeed help each other by providing estimated supervision on unlabeled data.
  
\emph{Knee results:} On knee MRIs, our method improves segmentation scores over separately learned networks by about 1.2 and 0.5, and registration scores increase by about 3.1 and 3.0, when training with 5 and 10 manual segmentation respectively. Especially for the challenging cartilage structures, our joint learning boosts segmentation by 1.4 and 0.7, and registration by 5.5 and 5.2 for N=5 and N=10 respectively.

\emph{Brain results:} Dice scores for segmentation and registration increase by about 2.6 and 3.5 respectively for the cortical structures of the brain MRIs.

\emph{One-shot learning:} In the one-shot experiments on both datasets, reasonable segmentation performance is achieved; moreover, DA increases the Dice score over unsupervised registration by about 2.7 and 1.8 on the knee and brain data respectively. This demonstrates the effectiveness of our framework for one-shot learning.

\begin{wraptable}{r}{0.5\textwidth}
\footnotesize
\vspace{-5mm}
\begin{adjustbox}{max width=0.5\textwidth}
    	\begin{tabular}{c|c|c|c}
    	\hline
    	 $N$ & Models & Seg Dice (\%) & Reg Dice (\%) \\
    	\hline
    	0 & Mono & - & 54.75(2.37)\\
    	\hline
    	\multirow{2}{*}{1} 	& {Semi-DA} & 61.19(1.49) & - \\
    						& {DA} &  \textbf{61.22(1.40)}  & 56.54(2.32) \\
    	\hline
    	\multirow{3}{*}{21} 	
    	& Mono & 73.48(2.58) & 59.47(2.34) \\
    	& {Semi-DA} & 75.63(1.66) &  62.92(2.14) \\
    	& {DA} & \textbf{76.06(1.50)} & \textbf{62.92(2.13)} \\
    	\hline
    	65 & Mono & 81.31(1.21) & 63.25(2.07) \\
    	\hline
    	\end{tabular}
\end{adjustbox}
	\caption{Segmentation and registration performance on 3D brain MRIs. Average(Standard deviation) of Dice scores (\%) for 31 cortical regions. $N$ of 65 training images are manually labeled.}
	\label{table:mindboogle}
	\vspace{-15mm}
\end{wraptable}

  \emph{Qualitative results:} DA achieves more anatomically consistent registrations than the mono-networks on the knee (Fig.~\ref{fig:qualitative_results}) and Brain MRI samples (see supplementary material). 

\section{Conclusion}
\label{section:conclusion}
We presented our DeepAtlas framework for joint learning of segmentation and registration networks using only few images with manual segmentations. By introducing an anatomical similarity loss, the learned registrations are more anatomically consistent. Furthermore, the segmentation network is guided by a form of data augmentation provided via the registration network on unlabeled images. For both bone/cartilage structures in knee MRIs and cortical structures in brain MRIs, our approach shows large improvements over separately learned networks. When only given one manual segmentation, our method provides one-shot segmentation learning and greatly improves registration. This demonstrates that one network can benefit from imperfect supervision on unlabeled data provided by the other network. Our approach provides a general solution to the lack of manual segmentations when training segmentation and registration networks. For future work, introducing uncertainty measures for the segmentation and registration networks may help alleviate the effect of poor predictions of one network on the other. It would also be of interest to investigate multitask learning via layer sharing for the segmentation and registration networks. This may further improve performance and decrease model size.

\textbf{Acknowledgements:} Research reported in this publication
was supported by the National Institutes of Health (NIH)
and the National Science Foundation (NSF) under award
numbers NSF EECS1711776 and NIH 1R01AR072013.
The content is solely the responsibility of the authors and
does not necessarily represent the official views of the NIH
or the NSF.

%
%
%
\bibliographystyle{splncs04}
\bibliography{miccai2019}
\newpage
 \appendix
 \section{Supplementary material}
 \vspace{-8mm}
 \begin{figure}[htp]
 	\begin{center}
 		\includegraphics[width=\textwidth]{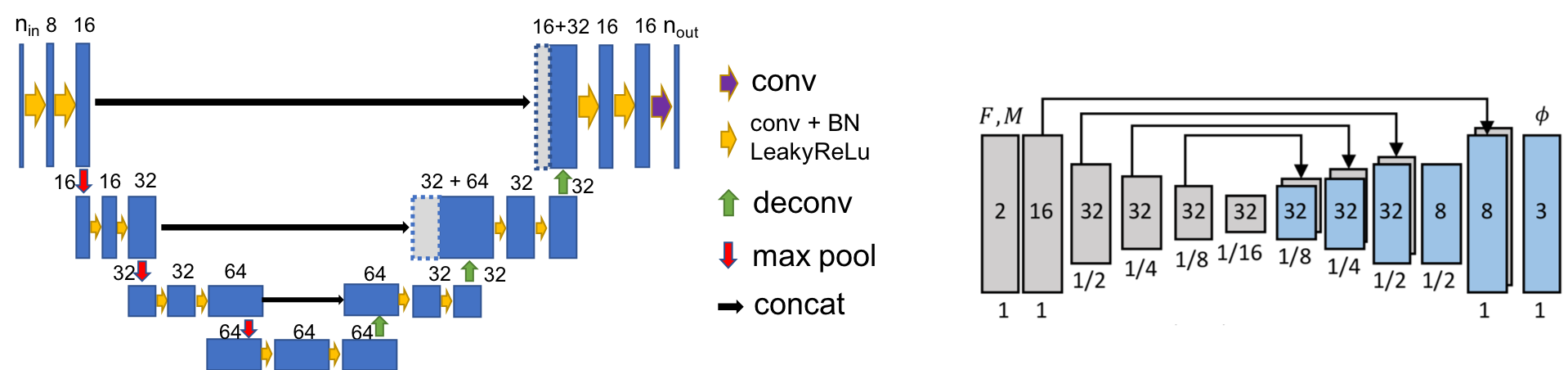}
 	\end{center}
 \caption{Architectures of the segmentation network (left) and the registration network~\cite{balakrishnan2018unsupervised} (right). In the segmentation network, max-pooling is used for down-sampling for which 2-stride convolution is used in the registration network.}
  \vspace{-5mm}
 \label{fig:networks}
 \end{figure}

\renewcommand{\arraystretch}{0}
\newcommand\cwidth{0.14\textwidth}
\begin{figure} [htp]
\begin{center}
\begin{adjustbox}{max width=\textwidth}
\begin{tabular}{ccccccc}
	Moving & Target &  Mono-0 & Mono-21 & DA-1 & DA-21 & Mono-65 {\smallskip}\\
\includegraphics[width=\cwidth]{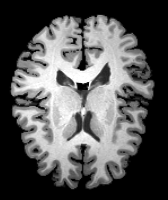} &
\includegraphics[width=\cwidth]{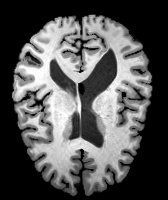} &
\includegraphics[width=\cwidth]{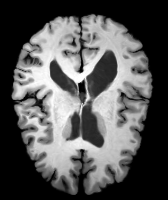} &
\includegraphics[width=\cwidth]{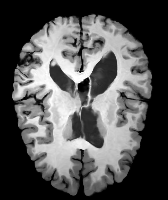} &
\includegraphics[width=\cwidth]{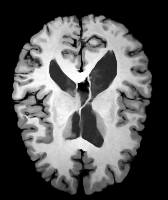} &
\includegraphics[width=\cwidth]{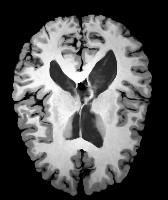} &
\includegraphics[width=\cwidth]{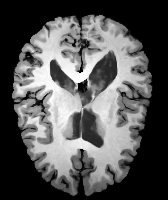} 
\\
\includegraphics[width=\cwidth]{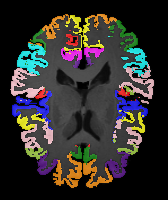} &
\includegraphics[width=\cwidth]{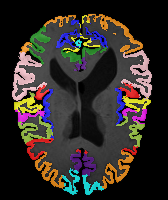} &
\includegraphics[width=\cwidth]{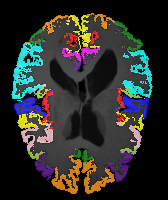} &
\includegraphics[width=\cwidth]{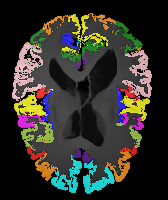} &
\includegraphics[width=\cwidth]{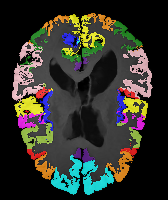} &
\includegraphics[width=\cwidth]{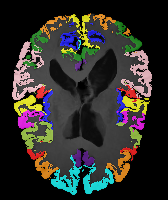} &
\includegraphics[width=\cwidth]{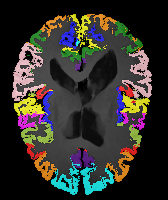} 
\\
\end{tabular}
\end{adjustbox}
\end{center}

\begin{center}
\renewcommand{\arraystretch}{0}
\newcommand\ccwidth{0.165\textwidth}
\begin{adjustbox}{max width=\textwidth}
\begin{tabular}{cccccc}
	Image & Manual Seg &  DA-1 & Mono-5 & DA-5 & Mono-200 {\smallskip}\\
\includegraphics[width=\ccwidth]{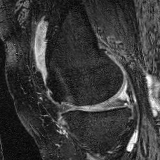} &
\includegraphics[width=\ccwidth]{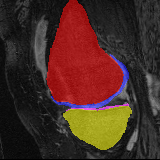} &
\includegraphics[width=\ccwidth]{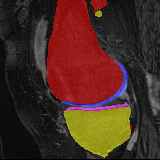} &
\includegraphics[width=\ccwidth]{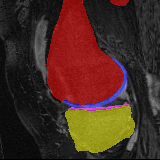} &
\includegraphics[width=\ccwidth]{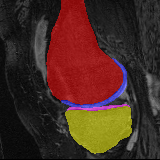} &
\includegraphics[width=\ccwidth]{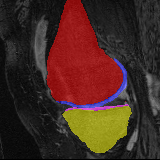} 
\\
\end{tabular}
\end{adjustbox}
\end{center}
\caption{€€Examples of brain MRI registration (top) and knee MRI segmentation (bottom) results. \textbf{Top}: The first two columns are the moving image/segmentation and the target image/segmentation followed by the warped moving images/segmentations by different models. \textbf{Bottom left to right}: original image, manual segmentation, and predictions of various models. Mono-$i$ and DA-$i$ represent the mono- and DA models trained with $i$ manual segmentations respectively.}
\vspace{-5mm}
\label{fig:qualitative_results_2}
\end{figure}

\end{document}